\documentclass{article}
\usepackage{ijcai23}

\usepackage{times}
\usepackage{soul}
\usepackage{url}
\usepackage[hidelinks]{hyperref}
\usepackage[utf8]{inputenc}
\usepackage[small]{caption}
\usepackage{graphicx}
\usepackage{amsmath}
\usepackage{amsthm}
\usepackage{amsfonts}
\usepackage{booktabs}
\usepackage{algorithm}
\usepackage{algorithmic}
\usepackage[switch]{lineno}
\usepackage{enumerate}
\usepackage[symbol]{footmisc}

\title{HOI-aware Adaptive Network for Weakly-supervised Action Segmentation}

\author{
Runzhong Zhang$^1$
\and
Suchen Wang$^1$\and
Yueqi Duan$^2$\footnote{Corresponding author}\and
Yansong Tang$^2$\and
\\
Yue Zhang$^3$\And
Yap-Peng Tan$^1$
\affiliations
$^1$Nanyang Technological University\\
$^2$Tsinghua University\\
$^3$Beijing Jiaotong University
\emails
\{runzhong001, suchen001\}@e.ntu.edu.sg,
duanyueqi@tsinghua.edu.cn,
tang.yansong@sz.tsinghua.edu.cn,
17112065@bjtu.edu.cn,
eyptan@ntu.edu.sg
}

\begin{document}

\maketitle

\begin{abstract}
In this paper, we propose an HOI-aware adaptive network named AdaAct for weakly-supervised action segmentation. Most existing methods learn a fixed network to predict the action of each frame with the neighboring frames. However, this would result in ambiguity when estimating similar actions, such as pouring juice and pouring coffee. To address this, we aim to exploit temporally global but spatially local human-object interactions (HOI) as video-level prior knowledge for action segmentation. The long-term HOI sequence provides crucial contextual information to distinguish ambiguous actions, where our network dynamically adapts to the given HOI sequence at test time. More specifically, we first design a video HOI encoder that extracts, selects, and integrates the most representative HOI throughout the video. Then, we propose a two-branch HyperNetwork to learn an adaptive temporal encoder, which automatically adjusts the parameters based on the HOI information of various videos on the fly. Extensive experiments on two widely-used datasets including Breakfast and 50Salads demonstrate the effectiveness of our method under different evaluation metrics.
\end{abstract}

\section{Introduction}

\begin{figure*}[t]
 \centering
 \includegraphics[width=1.0\linewidth]{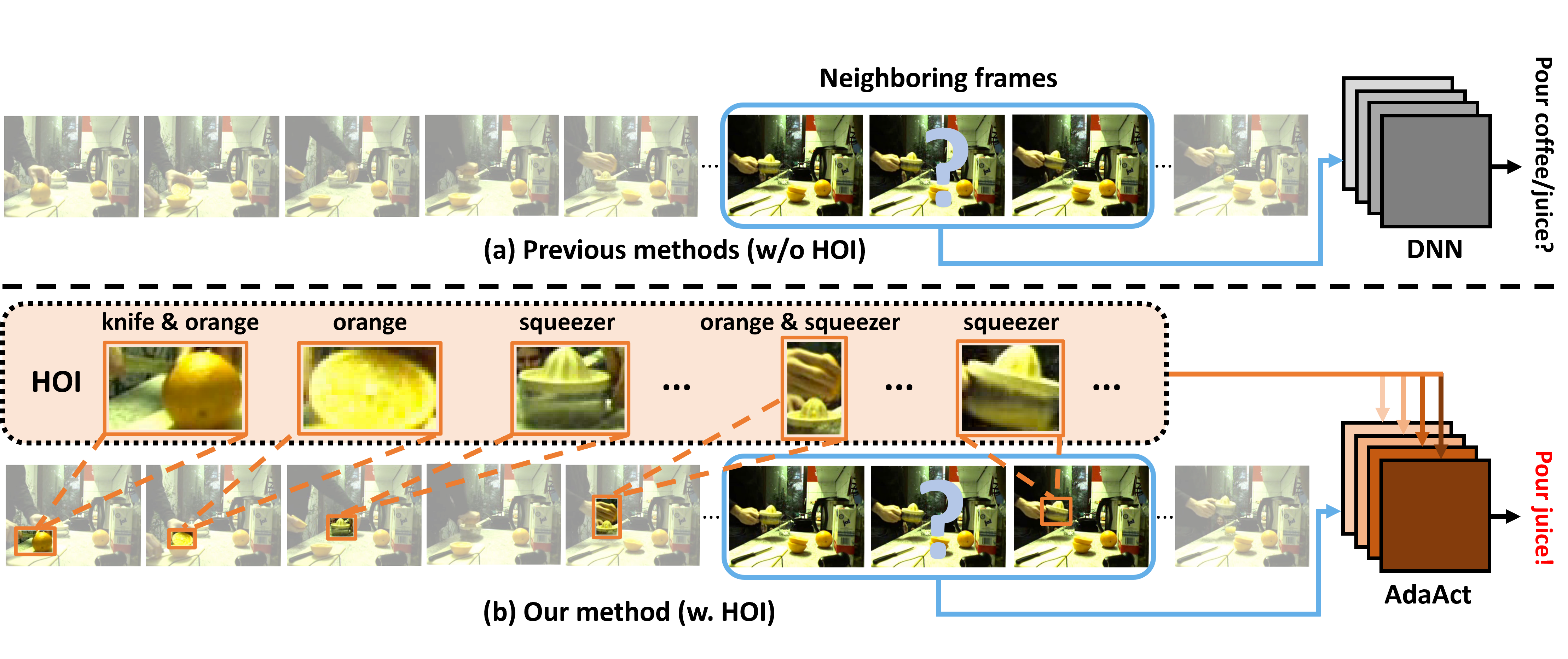}
  \caption{(a) Most existing methods estimate the action probability of frame $t$ using features of the adjacent frames (blue box in the figure). They are difficult to distinguish actions such as pouring juice, coffee, and water considering the high representation similarity. (b) Our method exploits temporally global but spatially local HOI information to learn an adaptive temporal encoder, which provides essential contextual information to distinguish similar actions. As shown in the figure, considering the interactions with a knife, orange, and squeezer in the video, the query action would be more likely to be pouring juice rather than coffee or water.}
  \label{fig:teaser}
\end{figure*}

Action segmentation aims to predict the action for every frame in the video. While previous methods have achieved remarkable performance in the fully-supervised setting~\cite{kuehne2016end,lea2017temporal,rohrbach2012database,singh2016multi,yi2021asformer,park2022maximization}, framewise annotation still requires huge labor costs and is hard to obtain. Therefore, action segmentation with weaker forms of supervision gradually gains its popularity in recent years. In particular, transcript supervision~\cite{bojanowski2014weakly,kuehne2017weakly,huang2016connectionist,ding2018weakly,li2019weakly,lu2021weakly} provides an ordered list of actions occurring in the video without the starting and ending time, which significantly reduces the annotation costs and improves the applicability to a rapidly-growing number of videos on the Internet.

To learn from the transcript, previous approaches mainly follow the ``generating-matching'' pipeline~\cite{richard2018neuralnetwork,li2019weakly,lu2021weakly}. With the given training videos, they first apply a temporal encoder to generate framewise action probabilities, and then match the predicted probabilities sequence with the transcript based on Viterbi decoding or dynamic time warping. However, to estimate the action probability of frame $t$ during the generating step, most existing approaches only take a fixed number of neighbor frames around it~\cite{richard2018neuralnetwork,li2019weakly}, and feed such video clip features into an RNN-based~\cite{chung2014empirical} architecture. In this case, the temporal encoder would fail to distinguish the attribute of similar actions such as pouring coffee and pouring juice, which may lead to counter-intuitive results of pouring coffee in a juice-making video. Although tremendous efforts have been made to remedy such ambiguity in the matching step, the results are still unsatisfying due to the inherent defect in the previous generating process.

In this paper, we address the ambiguity problem by designing an adaptive weakly-supervised action segmentation framework called AdaAct. Different from previous methods which take a series of fixed-length video clips as input successively (as shown in Figure~\ref{fig:teaser} (a)), we exploit rich contextual information from temporally global but spatially local human-object interactions (HOI) throughout the whole video. Such HOI sequence further instructs the network as prior knowledge, where our temporal encoder can be dynamically adapted to it at the test time. As illustrated in Figure~\ref{fig:teaser} (b), our method first extracts key interactions with objects at different video timestamps, such as the knife, orange, and squeezer. The obtained HOI sequence is further incorporated into the temporal encoder, thereby the network parameters of the encoder dynamically change with the HOI information on the fly. More specifically, we design a three-step video HOI encoder with the ``extracting-selecting-integrating'' process. We first apply a pre-trained HOI detector to extract positive interaction bounding boxes for the whole video and design a simple selecting algorithm to pick the most representative ones from them. Then, we explore the relations between these key HOI boxes and integrate them into a single feature vector via a transformer-based network. To dynamically adapt the network parameters, we propose a two-branch HyperNetwork that simultaneously learns HOI-dependent and HOI-independent knowledge. Our HOI-independent branch aims to unearth the general characteristics of instructional videos by iteratively updating a learnable embedding list throughout the training process. Such transferable information will later be encoded as a part of instruction used in the temporal encoder during the test time. On the other hand, our HOI-dependent branch takes the encoded feature vector from the video HOI detector as input and adapts the temporal encoder to the given HOI knowledge occurring in the video. Finally, late fusion is utilized to merge the knowledge from two branches, resulting in more precise action segmentation.

We summarize our key contributions as follows:

\begin{enumerate}[1)]
\item To our best knowledge, this is the first work to learn an adaptive temporal encoder for weakly-supervised action segmentation, where the parameters of the network are dynamically adapted according to the input video on the fly.

\item We propose to exploit temporally global but spatially local HOI information in weakly-supervised action segmentation, which provides essential contextual information to address the ambiguity problem of similar actions.

\item We validate our method on two challenging datasets, Breakfast and 50Salads, and achieve state-of-the-art results for both weakly-supervised action segmentation and alignment tasks.
\end{enumerate}

\section{Related Work}

\begin{figure*}[t]
 \centering
 \includegraphics[width=1.0\linewidth]{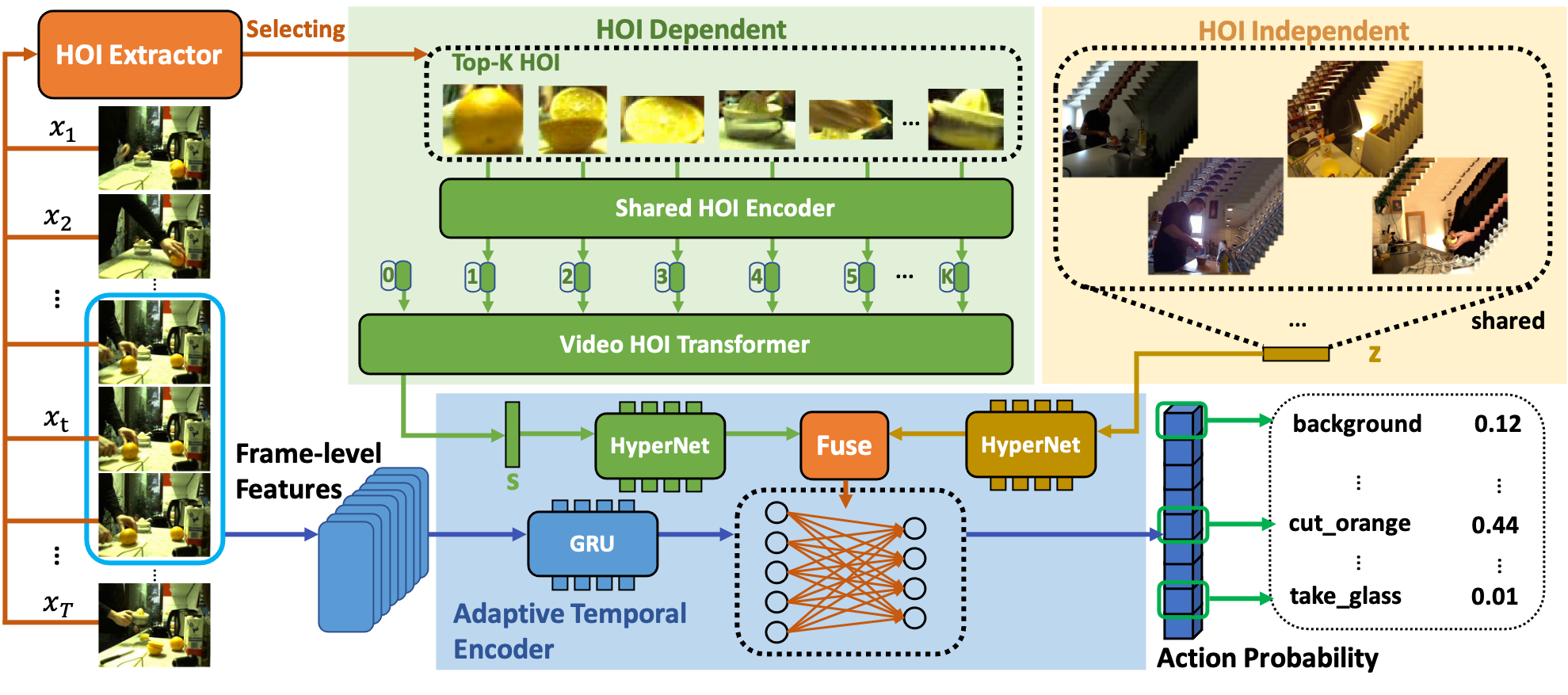}
  \caption{Overview of the network architecture. Our method simultaneously learns HOI-dependent knowledge $s$ from the video HOI encoder and HOI-independent knowledge $z$ from various videos across the dataset. The obtained knowledge is further incorporated through the two-branch HyperNetwork and late fusion, which generates the network parameters of the adaptive temporal encoder. In this way, our network dynamically adapts to the video contents when estimating the action probability of frame $t$, leading to better discrimination for similar actions.}
  \label{fig:network}
\end{figure*}

\paragraph{Fully-supervised action segmentation.} Fully-supervised action segmentation methods learn action segments under the guidance of framewise annotations. Earlier attempts~\cite{karaman2014fast,rohrbach2012database} applied action classification on the sliding window, followed by non-maximum suppression to filter out redundant predictions. However, these approaches failed to model the temporal dependency between action sequences. Kuehne \textit{et al}.~\shortcite{kuehne2016end} tackled this problem via the hidden Markov model, while Pirsiavash \textit{et al}.~\shortcite{pirsiavash2014parsing} applied context-free grammar to capture temporal structure. In recent years, various network architectures were proposed for learning long-range dependency. Lea \textit{et al}.~\shortcite{lea2017temporal} introduced an encoder-decoder architecture for action segmentation and detection. Lei \textit{et al}.~\shortcite{lei2018temporal} further applied deformable convolutions and residual stream. Farruha \textit{et al}.~\shortcite{farha2019ms} and Li \textit{et al}.~\shortcite{li2020ms} introduced dilated temporal convolution and proposed a multi-stage temporal convolutional network, while various methods improved the multi-stage network using graph-based temporal reasoning~\cite{huang2020improving} or boundary-aware cascade network~\cite{wang2020boundary}. With the success of transformer-based models in computer vision, Yi \textit{et al}.~\shortcite{yi2021asformer} first introduced the transformer into the action segmentation task. Different from previous methods, Li \textit{et al}.~\shortcite{li2022bridge} reformulated the action labels as text prompts and paired them with corresponding video clips, and co-trained the text encoder and the video encoder through a contrastive approach.

\paragraph{Weakly-supervised action segmentation.} Many of the weakly-supervised methods utilize transcripts as supervision during training. Huang \textit{et al}.~\shortcite{huang2016connectionist} first introduced the connectionist temporal classification framework to evaluate all possible matching between the videos and transcripts. Ding \textit{et al}.~\shortcite{ding2018weakly} started from the initial uniform mapping of the action transcript, and iteratively refined the transcript during the training procedure. However, these methods fail to achieve end-to-end training. Richard \textit{et al}.~\shortcite{richard2018neuralnetwork} instead generated pseudo frame labels using the Viterbi algorithm and trained a classifier based on framewise cross-entropy loss. Li \textit{et al}.~\shortcite{li2019weakly} further extended the NN-Viterbi~\cite{richard2018neuralnetwork} by introducing a new constrained discriminative forward loss, which maximized the energy difference between valid and invalid segmentation of training videos. In D3TW, Chang \textit{et al}.~\shortcite{chang2019d3tw} first applied a discriminative model for solving the degenerate sequence problem. As these methods have to search all the transcripts during testing and thus suffer from long inference time, Souri \textit{et al}.~\shortcite{souri2021fast} proposed MuCon, a two-branch network that predicted both transcript and framewise label of action segmentation, and designed the mutual loss to ensure the consistency of representations. In recent years, different weakly-supervised settings besides the transcripts have been studied. Fayyaz \textit{et al}.~\shortcite{fayyaz2020sct} and Li \textit{et al}.~\shortcite{li2020set} reduced the supervision level, assuming only the unordered list of actions is available for each training video. Inspired by the point supervision in semantic segmentation~\cite{bearman2016s}, Li \textit{et al}.~\shortcite{li2021temporal} trained a segmentation model using timestamps annotations, in which case only one arbitrary frame is annotated for each action. As these methods use different kinds of supervision for training, we do not directly compare them with our approach.

\paragraph{Human object interaction.} The existing HOI detection can be mainly categorized into single-stage approaches~\cite{liao2020ppdm,wang2020learning,kim2020uniondet,chen2021reformulating} and two-stage approaches~\cite{li2019transferable,zhang2021spatially,zhou2019relation,zhou2020cascaded,ulutan2020vsgnet}. Single-stage approaches integrate bounding boxes detection and interaction recognition into a single model. Liao \textit{et al}.~\shortcite{liao2020ppdm} and Wang \textit{et al}.~\shortcite{wang2020learning} first simultaneously generated bounding box candidates and interactions, and then outputted final predictions after the matching step. Chen \textit{et al}.~\shortcite{chen2021reformulating} instead reformed the HOI detection as an adaptive set prediction problem. Compared with one-stage methods, two-stage approaches first detect humans and objects following the object detection pipeline and then apply an interaction model to analyze the relations of the bounding boxes. Qi \textit{et al}.~\shortcite{qi2018learning} and Zhang \textit{et al}.~\shortcite{zhang2021spatially} modeled the relations using graph neural network. Fang \textit{et al}.~\shortcite{fang2018pairwise} emphasized the importance of human-part knowledge in HOI detection. Although different methods have been proposed in the image domain, research on video-level HOI detection is still under-exploited.

\section{Methodology}
Our goal is to address the weakly-supervised action segmentation problem under transcript supervision. Formally, we define each video with its supervision as a tuple $\left\{v, x^T_1, a^O_1, l^O_1\right\}$, where $v$ represents the video as a stack of raw frames, $x^T_1=[x_1, ..., x_T]$ denotes the unsupervised framewise features with length $T$, $a^O_1=[a_1, ..., a_O]$ indicates the transcript, an ordered list of $O$ actions occurred in the video, and $l^O_1=[l_1, ..., l_O]$ records the number of frames for each of the corresponding actions. Every action $a_o$ belongs to the set of $A$ action classes, namely $a_o\in\mathcal{A}=\left\{1,...,A\right\}$. During the inference, the objective is to predict the optimal action list $\hat{a}$ and corresponding length $\hat{l}$ based on the framewise features $x$ of the video $v$.

In this paper, we propose an adaptive network named AdaAct that utilizes video-level HOI to distinguish similar actions. As shown in Figure~\ref{fig:network}, our method mainly consists of a video HOI encoder and an adaptive temporal encoder. For the video HOI encoder, it first takes the input video and extracts all the valid interactions, then selects top-$K$ interactions by removing redundant and low-score detection. These interactions are finally integrated as HOI-dependent knowledge $s$. For the adaptive temporal encoder, it incorporates HOI-dependent knowledge $s$ with HOI-independent knowledge $z$ via a two-branch HOI-aware HyperNetwork~\cite{ha2016hypernetworks}, which predicts the network parameters of the temporal encoder. In the following, we describe the video HOI encoder and the adaptive temporal encoder in detail, as well as the training strategy to learn these two models.

\subsection{Video HOI Encoder}
The goal of our video HOI encoder is to model the dependencies between key HOI through the whole video and encode them as HOI-dependent knowledge $s$. It mainly contains three levels from the bottom to the top: extracting, selecting, and integrating.

\subsubsection{Extracting}
We take the video as input at the first level. Since the majority of HOI detection methods are developed only for image scenarios, we pre-process the video by down-sampling and extracting the raw frames under 15 FPS. After that, we employ the detector on every frame iteratively following the temporal order. To avoid introducing additional computation cost, here we follow the 100 Days of Hands~\cite{shan2020understanding} with its weight frozen during training and testing. The model outputs the predictions as tuple $\left\{b_h, b_o, c, t\right\}$, where $b_h$ and $b_o$ represent the bounding boxes of hands and object, $c\in [0, 1]$ denotes the interaction confidence score, and $t$ indicates the timestamp of the frame.

\subsubsection{Selecting}
Inspired by the non-maximum suppression (NMS)~\cite{neubeck2006efficient} used for filtering proposals in object detection, we propose a video-NMS algorithm to select top-$K$ object bounding boxes from the predictions pool. Different from the traditional NMS algorithm that filters the proposals only by the intersection over union (IoU), our method also adds the temporal constraint, so that the duplicates of highest-score proposals are removed based on IoU and time interval. After that, $K$ tuple predictions with the highest score are selected and ranked by timestamp order for the next step.

\subsubsection{Integrating}
In the integrating step, we propose a ViT-based network to generate HOI-dependent knowledge $s$. To handle all the $K$ object bounding boxes $b_o$, we use a frozen ResNet50~\cite{he2016deep} and project them into a sequence of HOI embeddings $[e_1, ..., e_K]$. Following ViT's design, we append a learnable embedding $e_{token}$ before the sequence, the state of which serves as the HOI-dependent knowledge $s$ at the transformer output. We also add the 1D learnable position embeddings $P=[p_{token}, p_1, ..., p_K]$ to the HOI embeddings and feed the resulting sequence into the ViT network.

Given the input $E_0 = [e_{token}, e_1, ... e_K] + P$, the network conducts the following procedures for layer $n$ from 1 to $N$:

\begin{align}
E'_n = \text{MSA}(\text{LN}(E_{n-1})) + E_{n-1}, \\
E_n = \text{MLP}(\text{LN}(E'_n)) + E'_n,
\end{align}
where MSA stands for the multi-head self-attention module, MLP represents multi-layer perceptron and LN denotes LayerNorm. The obtained HOI-dependent knowledge $s$ is then merged with HOI-independent knowledge $z$, which will be further explained in the following section.

\subsection{Adaptive Temporal Encoder}
For a fair comparison, we apply the GRU followed by a linear layer as the temporal encoder backbone in consistence with the previous methods~\cite{richard2018neuralnetwork,li2019weakly,lu2021weakly}. To instruct the temporal encoder with video-level knowledge, we employ the two-branch HOI-aware HyperNetwork ~\cite{ha2016hypernetworks}, a sub-network used to learn parameters for the temporal encoder in the action predicting process. Specifically, for the linear layer in the temporal encoder, its weights and bias are separately generated by feeding learnable embedding $z$ into the HOI-independent branch and $s$ into the HOI-dependent branch. The pipeline can be written as follows:

\begin{align}
W = F(H_i(z), H_d(s)), \\
b = F(H'_i(z), H'_d(s)),
\end{align}
where $H_i$, $H_d$ represent the independent and dependent branches for weights generation, and $H'_i$, $H'_d$ are for bias. Finally, we apply the late fusion module $F$ to integrate information from the two branches. Instead of fixing the network during test time in typical deep learning networks, our method adaptively adjusts the network parameters by incorporating different video-level prior knowledge into framewise action prediction, thus eliminating the potential ambiguity occurring between similar actions.

\subsubsection{Multi-head HOI-independent Branch}
Since the weight and bias can be considered as matrices with different dimensions, here we use the weight generation HyperNetwork as the example. We suppose the weight parameters generated from the HOI-independent branch are stored in matrix $W^z\in \mathbb{R}^{C_{out}\times A}$, where $C_{out}$ represents the frame representation dimension after processed by GRU. Therefore, the HOI-independent branch can be written as below:

\begin{equation}
  W^z = H_i(z).
\end{equation}

Instead of formulating the HOI-independent knowledge as a single vector, we initialize the embedding list $z=[z_1, ..., z_m]$, $z_i\in \mathbb{R}^D$. These vectors are fed into the two-layer linear network $H_i$, yielding $m$ different vectors with the same length $\frac{C_{out}\times A}{m}$. Finally, the outputs are reshaped and concatenated together as the $W^z$. To ensure the correctness of dimension, $C_{out}$ must be divisible by $m$. Formally, the network processes the following procedures:

\begin{align}
W^z_i = \varphi(\text{MLP}(z_i)), i=1, ..., m, \\
W = [W_1, ..., W_m], W_i\in \mathbb{R}^{\frac{C_{out}}{m} \times A},
\end{align}
where $\varphi(\cdot)$ represents the reshape operation. Compared with the original multi-head mechanism that uses different linear layers to project the same input, our method initializes a list of vectors and keeps the same network parameters.

\subsubsection{Multi-head HOI-dependent Branch}
Similar to the HOI-independent branch, we maintain the embedding list with the same size and separately sum them with the HOI-dependent knowledge $s$. The resulting vectors are projected by the two-layer linear network $H_d$, followed by the reshaping and concatenation to get the matrix $W^s$.

Finally, we generate the weight of the linear layer by element-wise multiplying $W^z$ and $W^s$:

\begin{equation}
  W = W^z \odot W^s.
\end{equation}

\subsection{Transcript Decoding and Training}

\begin{table}
    \centering
    \scalebox{0.85}{
    \begin{tabular}{l|cccc}
        \toprule
        \textbf{Breakfast} & MoF & MoF-BG & IoU & IoD\\
       \midrule
        ECTC~\cite{huang2016connectionist} &27.7 &- &- &-\\
        HMM/RNN~\cite{richard2017weakly} &33.3 &- &- &-\\
        TCFPN~\cite{ding2018weakly} &38.4 &38.4 &24.2 &40.6 \\
        NN-Viterbi*~\cite{richard2018neuralnetwork} &41.9 &38.9 &33.3 &42.8 \\
        D3TW~\cite{chang2019d3tw} &45.7 &- &- &- \\
        CDFL*~\cite{li2019weakly} &49.8 &47.1 &35.3 &45.6 \\
        MuCon~\cite{souri2021fast} &49.0 &- &- &- \\
        TASL*~\cite{lu2021weakly} &47.2 &44.4 &36.1 &45.8\\
        AdaAct (Ours) &\textbf{51.2} &\textbf{48.3} &\textbf{36.3} &\textbf{46.4} \\
        \bottomrule
        \toprule
        \textbf{50Salads} & MoF & MoF-BG & IoU & IoD\\
       \midrule
        NN-Viterbi~\cite{richard2018neuralnetwork} &49.4 &- &- &-\\
        CDFL~\cite{li2019weakly} &54.7 &49.8 &31.5 &40.4\\
        AdaAct (Ours) &\textbf{55.6} &\textbf{50.3} &\textbf{35.2} &\textbf{44.6} \\
        \bottomrule
    \end{tabular}}
    \caption{Action segmentation results on the Breakfast and the 50Salads datasets. The dash line indicates that no prior result is available. We report the reproduced results for the methods with an asterisk.}
    \label{tab:seg}
\end{table}

We formulate the action segmentation problem as finding the most likely labeling based on the video features. Specifically, the optimal $(\hat{a}_{1}^{O},\hat{l}_{1}^{O})$ can be obtained as follows:

\begin{equation}
\begin{split}
&\underset{a_{1}^{O}, l_{1}^{O}}{\arg \max }\left\{p\left(a_{1}^{O}, l_{1}^{O}\mid x_{1}^{T}\right)\right\} \\
&= \underset{a_{1}^{O}, l_{1}^{O}}{\arg \max }\left\{p\left(x_{1}^{T}\mid {a}_{1}^{O}, l_{1}^{O}\right) \cdot p\left(l_{1}^{O}\mid a_{1}^{O}\right) \cdot p\left(a_{1}^{O}\right)\right\}\\
&= \underset{a_{1}^{O}, l_{1}^{O}}{\arg \max }\left\{ \prod_{t=1}^{T} p\left(x_{t}\mid a_{o(t)}\right) 
\cdot \prod_{o=1}^{O} p\left(l_{o}\mid a_{o}\right) \cdot p\left(a_1^O\right)\right\}.
\end{split}
\end{equation}

In the above formula, $p\left(x_{t}\mid a\right)$ can be further transformed:

\begin{equation}
p\left(x_{t}\mid a\right) \propto \frac{p\left(a\mid x_{t}\right)}{p(a)},
  \label{eq:important}
\end{equation}
where $p\left(a\mid x_{t}\right)$ is modeled by the output of our adaptive temporal encoder. For the modeling of $p\left(l_{o}\mid a_{o}\right)$ and $p\left(a_1^O\right)$, the same settings with previous work~\cite{richard2018neuralnetwork} are utilized for the fair comparison. 

We apply the constrained discriminative forward loss proposed by ~\cite{li2019weakly} for the network training, and provide detailed comparisons with the baseline method in the following section. It is worth noting that our method shows great flexibility and can be plugged into different existing methods.

\begin{table}
    \centering
    \scalebox{0.85}{
    \begin{tabular}{l|cccc}
        \toprule
        \textbf{Breakfast} & MoF & MoF-BG & IoU & IoD\\
       \midrule
        ECTC~\cite{huang2016connectionist} &35.0 &- &- &45.0\\
        HMM/RNN~\cite{richard2017weakly} &- &- &- &47.3\\
        TCFPN~\cite{ding2018weakly} &53.5 &51.7 &35.3 &52.3 \\
        D3TW~\cite{chang2019d3tw} &57.0 &- &- &56.3 \\
        CDFL~\cite{li2019weakly} &63.0 &61.4 &45.8 &63.9 \\
        MuCon~\cite{souri2021fast} &- &- &- &\textbf{66.2} \\
        TASL~\cite{lu2021weakly} &64.1 &- &\textbf{49.9} &64.7\\
        AdaAct (Ours) &\textbf{64.4} &\textbf{62.3} &\textbf{49.9} &65.3 \\
        \bottomrule
        \toprule
        \textbf{50Salads} & MoF & MoF-BG & IoU & IoD\\
       \midrule
        CDFL~\cite{li2019weakly} &68.0 &65.3 &45.5 &58.7 \\
        AdaAct (Ours) &\textbf{69.8} &\textbf{66.5} &\textbf{47.5} &\textbf{60.3} \\
        \bottomrule
    \end{tabular}}
    \caption{Action alignment results on the Breakfast and the 50Salads datasets. The dash line indicates that no prior result is available.}
    \label{tab:align}
\end{table}

\begin{figure}[t]
 \centering
 \includegraphics[width=1.0\linewidth]{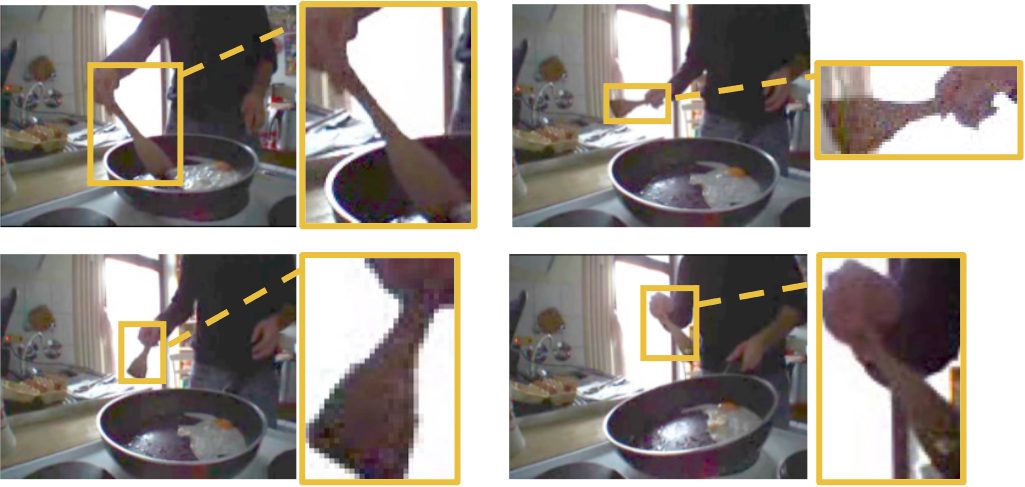}
  \caption{HOI detection in the frying egg activity. Our model only encodes the representation of spatula, since there is no direct interaction between human hands and egg.}
  \label{fig:egg}
\end{figure}

\section{Experiments}

\begin{table*}
    \centering
    \scalebox{0.81}{
    \begin{tabular}{l|cccccccccc|c}
        \toprule
         &cereals &coffee &fried-egg &juice &milk &pancake &salad &sandwich &scrambled-egg &tea &Total MoF\\
       \midrule
        NN-Viterbi~\cite{richard2018neuralnetwork} &39.5 &39.0 &48.4 &74.2 &56.8 &16.9 &46.0 &57.0 &45.0 &\textbf{44.9} &41.9\\
        CDFL~\cite{li2019weakly} &37.9 &37.0 &54.1 &75.8 &58.2 &31.1 &27.4 &38.8 &43.4 &34.6 &49.8\\
        TASL~\cite{lu2021weakly} &51.8 &43.6 &\textbf{59.2} &74.2 &56.5 &24.9 &46.0 &58.8 &\textbf{50.4} &42.1 &47.2\\
        AdaAct (Ours) &\textbf{56.1} &\textbf{57.3} &49.1 &\textbf{76.1} &\textbf{58.7} &\textbf{47.0} &\textbf{48.2} &\textbf{63.4} &44.7 &35.5 &\textbf{51.2}\\
        \bottomrule
    \end{tabular}}
    \caption{Action segmentation performance on the Breakfast dataset. We report the mean over frame accuracy (MoF) of every cooking activity and across all the activities, where ``cereals'' indicates ``making cereals'', etc.}
    \label{tab:allactions}
\end{table*}

\begin{figure*}[t]
 \centering
 \includegraphics[width=1.0\linewidth]{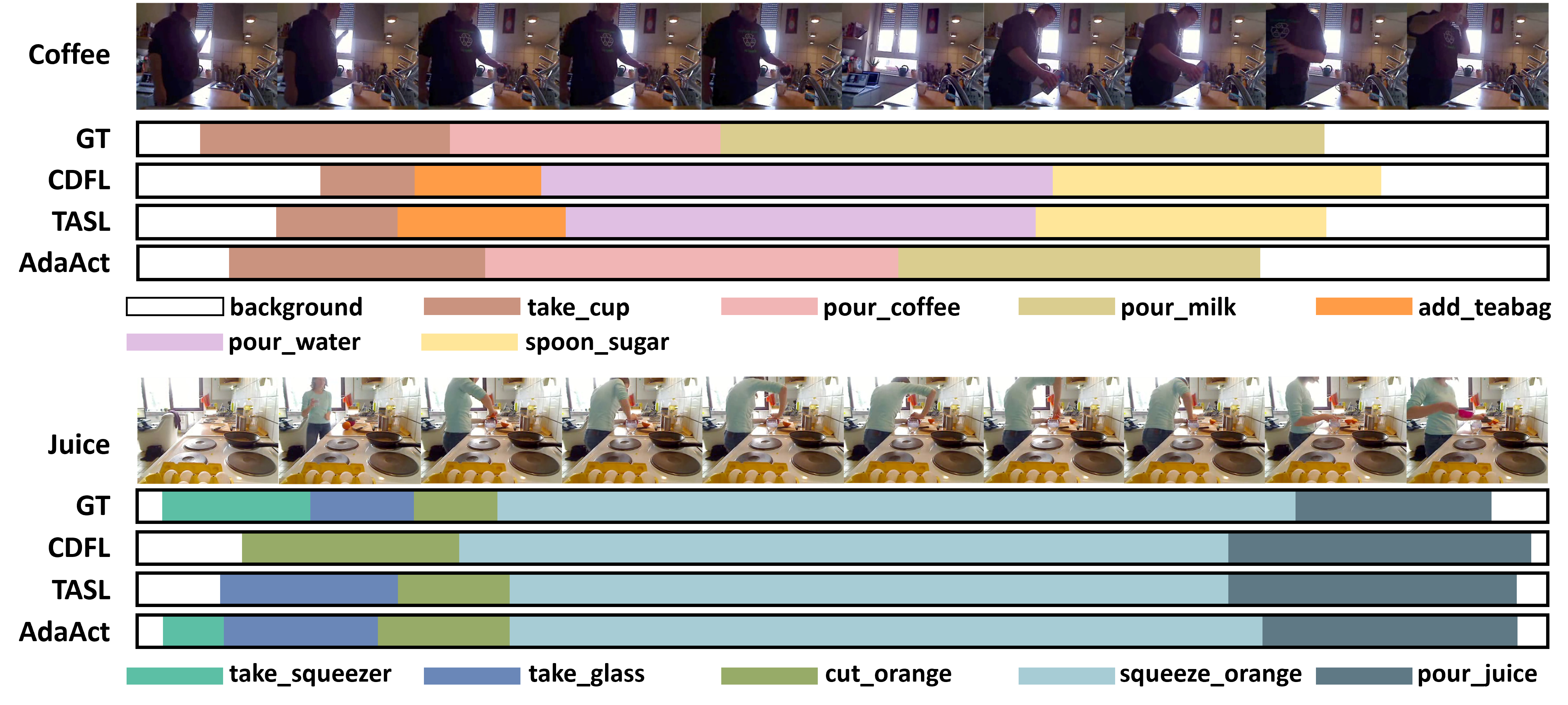}
  \caption{Action segmentation results of CDFL, TASL, and our approach on the coffee-making video (top) and juice-making video (bottom), where GT stands for the ground truth segmentation.}
  \label{fig:seg}
\end{figure*}

We validate our proposed method by comparing it with several state-of-the-art weakly-supervised action segmentation approaches, and discuss the effectiveness of each component in the following ablation studies.

\subsection{Experimental Setup}
\paragraph{Datasets.} We conduct our experiments on two real-world instructional video datasets: Breakfast~\cite{kuehne2014language} and 50Salads~\cite{stein2013combining}. The Breakfast dataset contains more than 1.7k videos of people performing 10 different cooking activities, such as preparing juice or preparing salad. The cooking activities are comprised of 48 fine-frained actions. Each video has 6.9 action segments on average, and the length of the video varies from several seconds to a few minutes. The 50Salads dataset has 50 long videos with 17 different action classes. On average, each video contains 20 action instances.

\paragraph{Evaluation metrics.} We use the following four metrics for evaluation. (1) Mean over frame accuracy (\textbf{MoF}) is defined as the number of correctly predicted frames divided by the total number of frames. (2) Mean over frame accuracy without background (\textbf{MoF-BG}) removes the background frames when calculating MoF, thus eliminating the drawback when video contains long periods of irrelevant information. (3) Intersection over union (\textbf{IoU}) is calculated as $\left | GT\cap correct \right | / \left | GT\cup  correct \right |$, where $GT$ stands for the ground truth frames and $correct$ denotes the correctly classified frames. (4) Intersection over detection (\textbf{IoD}) is defined as $\left | GT\cap correct \right | / \left | GT\right |$.

\paragraph{Implementation details.} For the video HOI encoder, we follow the same HOI detector pre-trained on the 100K dataset as mentioned in 100 Days of Hands~\cite{shan2020understanding}. We set 0.5 as the HOI detection threshold and pick $K=10$ bounding boxes after the selection process. For the ViT network, we replace the image patching and linear projection steps with the ResNet50 backbone, leading to $10\times 2048$ input size. We use $D=128$ for the dimension of both HOI-dependent and HOI-independent knowledge, and set the multi-head number as 8. For the adaptive temporal encoder, we use the 64-hidden unit GRU. We maintain the learning rate of 0.01 with 12500 epochs through the training process.

\subsection{Experimental Results}
We report the experimental results for two tasks, namely action segmentation where only the video is available during the inference, and action alignment where both video and transcript are provided.

\subsubsection{Quantitative Results}
We quantitatively compare our method with prior works in this section. Table~\ref{tab:seg} reports the action segmentation results on two instructional video datasets under four evaluation metrics, where the best results are indicated in bold. We can observe that by introducing HOI-aware knowledge, our method exceeds state-of-the-art methods by 1.4\% MoF and 1.2\% MoF-BG on the Breakfast dataset, and 0.9\% MoF and 0.5\% MoF-BG on the 50Salads dataset. This validates that when only video is given during testing, our method learns rich video-level knowledge and instructs the decision-making of the temporal encoder, leading to significant performance improvement on both datasets.

Table~\ref{tab:align} shows the action alignment results following the same metrics in Table~\ref{tab:seg}. Notice that in this setting the transcript is available during inference, thus providing stronger video-level knowledge compared with learned HOI-aware knowledge in our method. Despite this, our method still outperforms existing approaches on both datasets and achieves +1.8\% MoF and 1.2\% MoF-BG improvement on the 50Salads, which proves that the HOI-aware knowledge also helps to refine the starts and ends of predicted actions in the video.

\begin{figure}[t]
 \centering
 \includegraphics[width=1.0\linewidth]{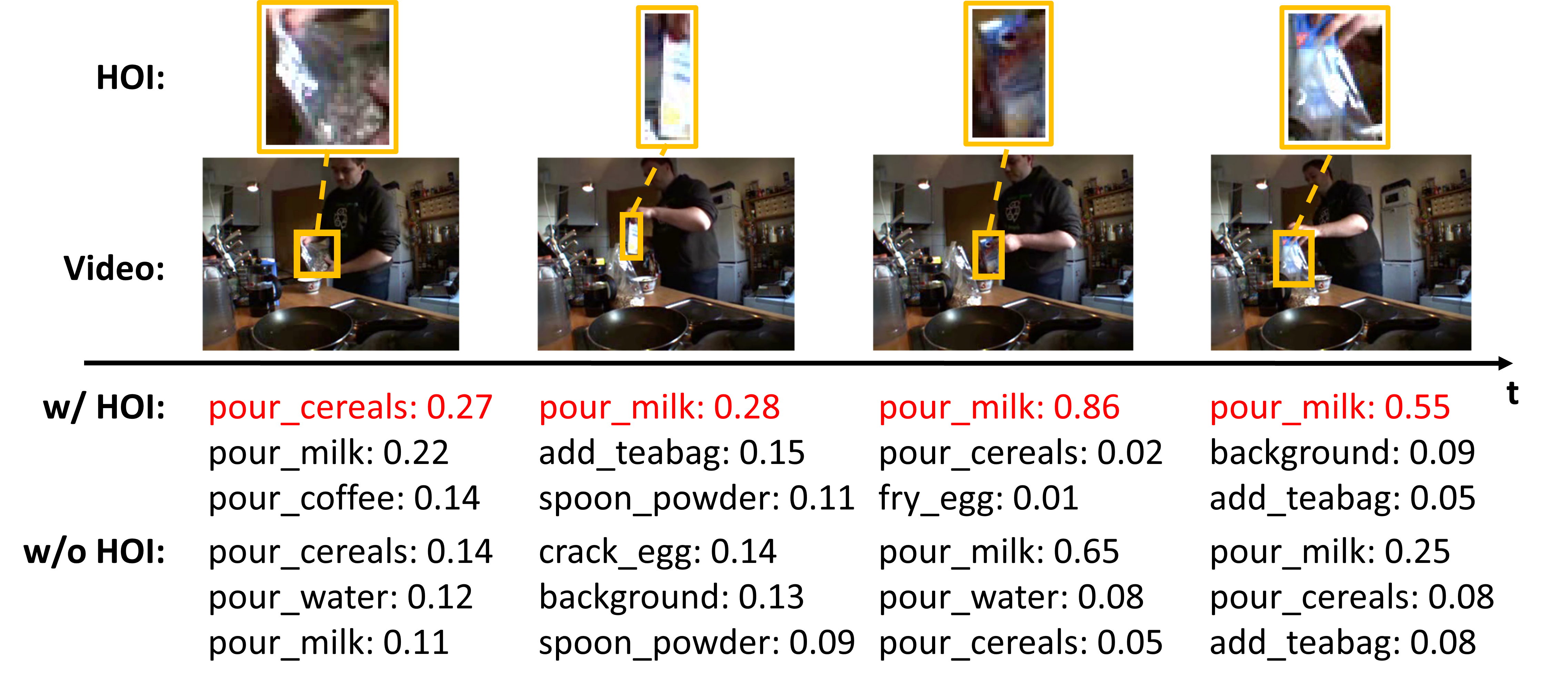}
  \caption{Visualization of HOI detection and corresponding action probability results. Only 4 frames with top-3 action predictions are shown due to the figure size limit. The correct predictions with the highest probability score are marked in red color.}
  \label{fig:prob}
\end{figure}

We also report the per-activity MoF results on the Breakfast dataset under the action segmentation setting. In Table~\ref{tab:allactions}, we can observe that our methods outperforms the baseline approaches for most of the cooking activities as expected. For those activities containing similar actions, such as pouring cereals in ``cereals'', pouring coffee in ``coffee'', and pouring milk in ``pancake'', our method achieves a large performance gain (+4.3\% MoF in ``cereals'', +13.7\% MoF in ``coffee'' and +15.9\% MoF in ``pancake''). This validates the effectiveness of our method in distinguishing ambiguous actions among different cooking activities.

However, we notice that our method still suffers from low performance in tea-making and egg-making (``fried-egg'' and ``scrambled-egg'') videos. Since our method applies the constrained discriminative forward loss and follows the temporal encoder architecture in CDFL, our method is inevitably affected by its performance. From this perspective, we still outperforms CDFL for +0.9\% in ``tea'' and +1.3\% in ``scrambled-egg''. We also investigate the reason for performance dropping in the ``fried-egg'' activity. By visualizing the selected HOI bounding boxes in Figure~\ref{fig:egg}, we observe that our HOI extractor tends to capture the spatula, while limited egg information is selected due to the long distance between itself and human hands. Despite the fact that people in the videos directly interact with the spatula for a long period of time, with the absence of egg detection, such HOI still could not provide enough information to instruct the temporal encoder. Therefore, the performance of our method instead degrades due to the noisy HOI-aware knowledge.

\subsubsection{Qualitative Results}
Figure~\ref{fig:seg} shows the action segmentation results of two videos on the Breakfast dataset. For the coffee-making video on the top, existing methods make the wrong predictions due to the high similarity of different pouring actions. In contrast, our method encodes strong semantic information in the HOI-aware knowledge, thus correctly classifying all the actions contained in the video. In the bottom juice-making case, when all the methods successfully capture the actions through the video, our method also shows higher accuracy in detecting the boundaries among different actions.

We also visualize how HOI detection helps to eliminate ambiguity in predicting action probability. For the cereals-making video in Figure~\ref{fig:prob}, our HOI detector precisely captures the interactions with a cereal bag and a milk box at different timestamps. Without applying HOI-aware knowledge, the existing method suffers from low confidence when distinguishing ``pour\_cereals'', ``pour\_water'' and ``pour\_milk'' in the first frame and makes the wrong prediction in the second. In contrast, our method both makes the correct predictions and widens the probability gaps among similar actions. In the last two frames, our method also greatly improves the ``pour\_milk'' confidence, demonstrating that HOI-aware knowledge provides strong instruction for a better action probability estimation.

\subsection{Ablation Studies}
We examine different components of our method and report the results in Table~\ref{tab:ablall}. The full model with the best performance is provided at the bottom for comparison. Introducing HOI-dependent knowledge leads to the most significant improvement of MoF by 3.7\% and HOI-independent knowledge contributes to 0.9\% MoF improvement, which demonstrates that both sources of knowledge are necessary for the HOI-aware understanding. In addition, applying the multi-head mechanism further achieves +1.8\% MoF.   

Table~\ref{tab:abldim} shows how different dimensions of HOI-dependent/independent knowledge affect the action segmentation. As expected, either too small or large size would cause the performance to drop. The highest accuracy is achieved with 128 dimensions. 

\section{Conclusion}
In this paper, we have proposed AdaAct, an HOI-aware adaptive network for video action segmentation under transcript supervision. Our method exploits essential contextual information from temporally global but spatially local human-object interactions, and dynamically adapts its network parameters according to the videos on the fly. AdaAct achieves state-of-the-art results on two instructional video datasets for both action segmentation and alignment tasks, and especially shows strong capability in distinguishing similar actions.

\begin{table}
    \centering
    \scalebox{0.95}{
    \begin{tabular}{ccc|c}
        \toprule
        HOI-dependent & HOI-independent & multi-head & MoF\\
       \midrule
         &\checkmark &\checkmark &47.5 \\
         \checkmark & &\checkmark &50.3 \\
         \checkmark &\checkmark & &49.4 \\
        \checkmark &\checkmark &\checkmark &\textbf{51.2} \\
        \bottomrule
    \end{tabular}}
    \caption{Effect of each component in our method. We report the mean over frame accuracy (MoF) for action segmentation on the Breakfast dataset.}
    \label{tab:ablall}
\end{table}

\begin{table}
    \centering
    \scalebox{0.95}{
    \begin{tabular}{c|cccc}
        \toprule
        knowledge dimension & MoF & MoF-BG & IoU & IoD\\
       \midrule
        32 &45.5 &42.8 &33.1 &42.9\\
        64 &48.7 &46.0 &34.8 &45.2\\
        128 &\textbf{51.2} &\textbf{48.3} &\textbf{36.3} &\textbf{46.4}\\
        256 &46.9 &44.0 &33.7 &44.0\\
        \bottomrule
    \end{tabular}}
    \caption{Effect of the HOI-dependent and HOI-independent knowledge dimension in our method. We report the mean over frame accuracy (MoF) for action segmentation on the Breakfast dataset.}
    \label{tab:abldim}
\end{table}

\section*{Acknowledgements}
This research is supported in part by the National Research Foundation of Singapore under the NRF Medium Sized Centre Scheme (CARTIN), and in part by the National Natural Science Foundation of China under Grant 62206147 and Grant 62206153. Any opinions, findings and conclusions expressed in this material are those of the author(s) and do not reflect the views of National Research Foundation, Singapore and National Natural Science Foundation, China. 

\bibliographystyle{named}
\bibliography{ijcai23}

@inproceedings{karaman2014fast,
  title={Fast saliency based pooling of fisher encoded dense trajectories},
  author={Karaman, Svebor and Seidenari, Lorenzo and Del Bimbo, Alberto},
  booktitle={ECCV THUMOS Workshop},
  volume={1},
  number={2},
  pages={5},
  year={2014}
}

@inproceedings{rohrbach2012database,
  title={A database for fine grained activity detection of cooking activities},
  author={Rohrbach, Marcus and Amin, Sikandar and Andriluka, Mykhaylo and Schiele, Bernt},
  booktitle={CVPR},
  pages={1194--1201},
  year={2012},
}

@inproceedings{kuehne2016end,
  title={An end-to-end generative framework for video segmentation and recognition},
  author={Kuehne, Hilde and Gall, Juergen and Serre, Thomas},
  booktitle={WACV},
  pages={1--8},
  year={2016},
}

@inproceedings{pirsiavash2014parsing,
  title={Parsing videos of actions with segmental grammars},
  author={Pirsiavash, Hamed and Ramanan, Deva},
  booktitle={CVPR},
  pages={612--619},
  year={2014}
}

@inproceedings{farha2019ms,
  title={Ms-tcn: Multi-stage temporal convolutional network for action segmentation},
  author={Farha, Yazan Abu and Gall, Jurgen},
  booktitle={CVPR},
  pages={3575--3584},
  year={2019}
}

@article{li2020ms,
  title={Ms-tcn++: Multi-stage temporal convolutional network for action segmentation},
  author={Li, Shi-Jie and AbuFarha, Yazan and Liu, Yun and Cheng, Ming-Ming and Gall, Juergen},
  journal={TPAMI},
  year={2020},
}

@inproceedings{wang2020boundary,
  title={Boundary-aware cascade networks for temporal action segmentation},
  author={Wang, Zhenzhi and Gao, Ziteng and Wang, Limin and Li, Zhifeng and Wu, Gangshan},
  booktitle={ECCV},
  pages={34--51},
  year={2020},
}

@inproceedings{huang2020improving,
  title={Improving action segmentation via graph-based temporal reasoning},
  author={Huang, Yifei and Sugano, Yusuke and Sato, Yoichi},
  booktitle={CVPR},
  pages={14024--14034},
  year={2020}
}

@inproceedings{lea2017temporal,
  title={Temporal convolutional networks for action segmentation and detection},
  author={Lea, Colin and Flynn, Michael D and Vidal, Rene and Reiter, Austin and Hager, Gregory D},
  booktitle={CVPR},
  pages={156--165},
  year={2017}
}

@inproceedings{lei2018temporal,
  title={Temporal deformable residual networks for action segmentation in videos},
  author={Lei, Peng and Todorovic, Sinisa},
  booktitle={CVPR},
  pages={6742--6751},
  year={2018}
}

@article{yi2021asformer,
  title={Asformer: Transformer for action segmentation},
  author={Yi, Fangqiu and Wen, Hongyu and Jiang, Tingting},
  journal={arXiv preprint arXiv:2110.08568},
  year={2021}
}

@inproceedings{richard2018neuralnetwork,
  title={Neuralnetwork-viterbi: A framework for weakly supervised video learning},
  author={Richard, Alexander and Kuehne, Hilde and Iqbal, Ahsan and Gall, Juergen},
  booktitle={CVPR},
  pages={7386--7395},
  year={2018}
}

@inproceedings{li2019weakly,
  title={Weakly supervised energy-based learning for action segmentation},
  author={Li, Jun and Lei, Peng and Todorovic, Sinisa},
  booktitle={ICCV},
  pages={6243--6251},
  year={2019}
}

@article{souri2021fast,
  title={Fast weakly supervised action segmentation using mutual consistency},
  author={Souri, Yaser and Fayyaz, Mohsen and Minciullo, Luca and Francesca, Gianpiero and Gall, Juergen},
  journal={TPAMI},
  year={2021}
}

@inproceedings{huang2016connectionist,
  title={Connectionist temporal modeling for weakly supervised action labeling},
  author={Huang, De-An and Fei-Fei, Li and Niebles, Juan Carlos},
  booktitle={ECCV},
  pages={137--153},
  year={2016}
}

@inproceedings{chang2019d3tw,
  title={D3tw: Discriminative differentiable dynamic time warping for weakly supervised action alignment and segmentation},
  author={Chang, Chien-Yi and Huang, De-An and Sui, Yanan and Fei-Fei, Li and Niebles, Juan Carlos},
  booktitle={CVPR},
  pages={3546--3555},
  year={2019}
}

@inproceedings{ding2018weakly,
  title={Weakly-supervised action segmentation with iterative soft boundary assignment},
  author={Ding, Li and Xu, Chenliang},
  booktitle={CVPR},
  pages={6508--6516},
  year={2018}
}

@inproceedings{fayyaz2020sct,
  title={Sct: Set constrained temporal transformer for set supervised action segmentation},
  author={Fayyaz, Mohsen and Gall, Jurgen},
  booktitle={CVPR},
  pages={501--510},
  year={2020}
}

@inproceedings{li2020set,
  title={Set-constrained viterbi for set-supervised action segmentation},
  author={Li, Jun and Todorovic, Sinisa},
  booktitle={CVPR},
  pages={10820--10829},
  year={2020}
}

@inproceedings{bearman2016s,
  title={What’s the point: Semantic segmentation with point supervision},
  author={Bearman, Amy and Russakovsky, Olga and Ferrari, Vittorio and Fei-Fei, Li},
  booktitle={ECCV},
  pages={549--565},
  year={2016}
}

@inproceedings{li2021temporal,
  title={Temporal Action Segmentation from Timestamp Supervision},
  author={Li, Zhe and Abu Farha, Yazan and Gall, Jurgen},
  booktitle={CVPR},
  pages={8365--8374},
  year={2021}
}

@inproceedings{singh2016multi,
  title={A multi-stream bi-directional recurrent neural network for fine-grained action detection},
  author={Singh, Bharat and Marks, Tim K and Jones, Michael and Tuzel, Oncel and Shao, Ming},
  booktitle={CVPR},
  pages={1961--1970},
  year={2016}
}

@article{park2022maximization,
  title={Maximization and restoration: Action segmentation through dilation passing and temporal reconstruction},
  author={Park, Junyong and Kim, Daekyum and Huh, Sejoon and Jo, Sungho},
  journal={Pattern Recognition},
  volume={129},
  pages={108764},
  year={2022},
}

@inproceedings{lu2021weakly,
  title={Weakly-supervised action segmentation and alignment via transcript-aware union-of-subspaces learning},
  author={Lu, Zijia and Elhamifar, Ehsan},
  booktitle={ICCV},
  pages={8085--8095},
  year={2021}
}

@inproceedings{bojanowski2014weakly,
  title={Weakly supervised action labeling in videos under ordering constraints},
  author={Bojanowski, Piotr and Lajugie, R{\'e}mi and Bach, Francis and Laptev, Ivan and Ponce, Jean and Schmid, Cordelia and Sivic, Josef},
  booktitle={ECCV},
  pages={628--643},
  year={2014}
}

@article{kuehne2017weakly,
  title={Weakly supervised learning of actions from transcripts},
  author={Kuehne, Hilde and Richard, Alexander and Gall, Juergen},
  journal={Computer Vision and Image Understanding},
  volume={163},
  pages={78--89},
  year={2017},
  publisher={Elsevier}
}

@article{chung2014empirical,
  title={Empirical evaluation of gated recurrent neural networks on sequence modeling},
  author={Chung, Junyoung and Gulcehre, Caglar and Cho, KyungHyun and Bengio, Yoshua},
  journal={arXiv preprint arXiv:1412.3555},
  year={2014}
}

@inproceedings{shan2020understanding,
  title={Understanding human hands in contact at internet scale},
  author={Shan, Dandan and Geng, Jiaqi and Shu, Michelle and Fouhey, David F},
  booktitle={CVPR},
  pages={9869--9878},
  year={2020}
}

@article{ha2016hypernetworks,
  title={Hypernetworks},
  author={Ha, David and Dai, Andrew and Le, Quoc V},
  journal={arXiv preprint arXiv:1609.09106},
  year={2016}
}

@inproceedings{liao2020ppdm,
  title={Ppdm: Parallel point detection and matching for real-time human-object interaction detection},
  author={Liao, Yue and Liu, Si and Wang, Fei and Chen, Yanjie and Qian, Chen and Feng, Jiashi},
  booktitle={CVPR},
  pages={482--490},
  year={2020}
}

@inproceedings{kim2020uniondet,
  title={Uniondet: Union-level detector towards real-time human-object interaction detection},
  author={Kim, Bumsoo and Choi, Taeho and Kang, Jaewoo and Kim, Hyunwoo J},
  booktitle={ECCV},
  pages={498--514},
  year={2020},
}

@inproceedings{li2019transferable,
  title={Transferable interactiveness knowledge for human-object interaction detection},
  author={Li, Yong-Lu and Zhou, Siyuan and Huang, Xijie and Xu, Liang and Ma, Ze and Fang, Hao-Shu and Wang, Yanfeng and Lu, Cewu},
  booktitle={CVPR},
  pages={3585--3594},
  year={2019}
}

@inproceedings{zhou2019relation,
  title={Relation parsing neural network for human-object interaction detection},
  author={Zhou, Penghao and Chi, Mingmin},
  booktitle={ICCV},
  pages={843--851},
  year={2019}
}

@inproceedings{ulutan2020vsgnet,
  title={Vsgnet: Spatial attention network for detecting human object interactions using graph convolutions},
  author={Ulutan, Oytun and Iftekhar, ASM and Manjunath, Bangalore S},
  booktitle={CVPR},
  pages={13617--13626},
  year={2020}
}

@inproceedings{zhou2020cascaded,
  title={Cascaded human-object interaction recognition},
  author={Zhou, Tianfei and Wang, Wenguan and Qi, Siyuan and Ling, Haibin and Shen, Jianbing},
  booktitle={CVPR},
  pages={4263--4272},
  year={2020}
}

@inproceedings{wang2020learning,
  title={Learning human-object interaction detection using interaction points},
  author={Wang, Tiancai and Yang, Tong and Danelljan, Martin and Khan, Fahad Shahbaz and Zhang, Xiangyu and Sun, Jian},
  booktitle={CVPR},
  pages={4116--4125},
  year={2020}
}

@inproceedings{chen2021reformulating,
  title={Reformulating hoi detection as adaptive set prediction},
  author={Chen, Mingfei and Liao, Yue and Liu, Si and Chen, Zhiyuan and Wang, Fei and Qian, Chen},
  booktitle={CVPR},
  pages={9004--9013},
  year={2021}
}

@inproceedings{qi2018learning,
  title={Learning human-object interactions by graph parsing neural networks},
  author={Qi, Siyuan and Wang, Wenguan and Jia, Baoxiong and Shen, Jianbing and Zhu, Song-Chun},
  booktitle={ECCV},
  pages={401--417},
  year={2018}
}

@inproceedings{zhang2021spatially,
  title={Spatially conditioned graphs for detecting human-object interactions},
  author={Zhang, Frederic Z and Campbell, Dylan and Gould, Stephen},
  booktitle={ICCV},
  pages={13319--13327},
  year={2021}
}

@inproceedings{fang2018pairwise,
  title={Pairwise body-part attention for recognizing human-object interactions},
  author={Fang, Hao-Shu and Cao, Jinkun and Tai, Yu-Wing and Lu, Cewu},
  booktitle={ECCV},
  pages={51--67},
  year={2018}
}

@inproceedings{neubeck2006efficient,
  title={Efficient non-maximum suppression},
  author={Neubeck, Alexander and Van Gool, Luc},
  booktitle={ICPR},
  volume={3},
  pages={850--855},
  year={2006},
}

@inproceedings{he2016deep,
  title={Deep residual learning for image recognition},
  author={He, Kaiming and Zhang, Xiangyu and Ren, Shaoqing and Sun, Jian},
  booktitle={CVPR},
  pages={770--778},
  year={2016}
}

@inproceedings{kuehne2014language,
  title={The language of actions: Recovering the syntax and semantics of goal-directed human activities},
  author={Kuehne, Hilde and Arslan, Ali and Serre, Thomas},
  booktitle={CVPR},
  pages={780--787},
  year={2014}
}

@inproceedings{richard2017weakly,
  title={Weakly supervised action learning with rnn based fine-to-coarse modeling},
  author={Richard, Alexander and Kuehne, Hilde and Gall, Juergen},
  booktitle={CVPR},
  pages={754--763},
  year={2017}
}

@inproceedings{stein2013combining,
  title={Combining embedded accelerometers with computer vision for recognizing food preparation activities},
  author={Stein, Sebastian and McKenna, Stephen J},
  booktitle={UbiComp},
  pages={729--738},
  year={2013}
}

@inproceedings{li2022bridge,
  title={Bridge-prompt: Towards ordinal action understanding in instructional videos},
  author={Li, Muheng and Chen, Lei and Duan, Yueqi and Hu, Zhilan and Feng, Jianjiang and Zhou, Jie and Lu, Jiwen},
  booktitle={CVPR},
  pages={19880--19889},
  year={2022}
}

\end{document}